\documentclass[10pt,twocolumn,letterpaper]{article}

\usepackage{wacv}
\usepackage{times}
\usepackage{epsfig}
\usepackage{graphicx}
\usepackage{amsmath}
\usepackage{amssymb}

\def\wacvPaperID{11} 

\wacvfinalcopy 

\ifwacvfinal
\def\assignedStartPage{1} 
\fi

\ifwacvfinal
\usepackage[breaklinks=true,bookmarks=false]{hyperref}
\else
\usepackage[pagebackref=true,breaklinks=true,colorlinks,bookmarks=false]{hyperref}
\fi

\ifwacvfinal
\else
\fi

\begin{document}

\title{Improved 2D Keypoint Detection in Out-of-Balance and Fall Situations – combining input rotations and a kinematic model}


\author{Michael Zwölfer\\
University of Innsbruck\\
{\tt\small michael.zwoelfer@uibk.ac.at}
\and
Dieter Heinrich\\
University of Innsbruck\\
{\tt\small dieter.heinrich@uibk.ac.at}
\and
Kurt Schindelwig\\University of Innsbruck\\
{\tt\small kurt.schindelwig@uibk.ac.at}
\and
Bastian Wandt\\
University of British Columbia\\
{\tt\small wandt@cs.ubc.ca}
\and
Helge Rhodin\\
University of British Columbia\\
{\tt\small rhodin@cs.ubc.ca}
\and
Jörg Spörri\\
University of Zurich\\
{\tt\small joerg.spoerri@balgrist.ch}
\and
Werner Nachbauer\\
University of Innsbruck\\
{\tt\small werner.nachbauer@uibk.ac.at}
}

\maketitle

\begin{abstract}
  Injury analysis may be one of the most beneficial applications of deep learning based human pose estimation. To facilitate further research on this topic, we provide an injury specific 2D dataset for alpine skiing, covering in total 533 images. We further propose a post processing routine, that combines rotational information with a simple kinematic model. We could improve detection results in fall situations by up to 21\% regarding the PCK@0.2 metric.
\end{abstract}

\section{Introduction}
Alpine ski racing is considered an extreme sport, exposing its athletes to a very high risk of injury \cite{barth}. Despite many scientific efforts and the introduction of various prevention measures, the incidence of ACL injuries among professional ski racers has been constantly growing over the last twenty years \cite{barth}. A quantitative biomechanical analysis of the underlying injury mechanisms is challenging and so far limited to time intensive and error prone manual matching methods \cite{bere}. Today, deep learning based keypoint detection serves in many real-world applications already. It provides a fast and convenient method for 2D human motion capture and has therefore become an important tool in biomechanical research and sports analysis \cite{intro}. Furthermore, it has the potential to overcome the limitations of manual matching and enable a comprehensive, qualitative analysis of injury situations that can not be measured with conventional motion capture methods, e.g. IMU sensors or multiple camera systems. In recent years, ready to use deep learning algorithms such as OpenPose \cite{openpose}, AlphaPose \cite{alphapose} or DCPose \cite{dcpose} received great interest. Although these algorithms provide accurate keypoint estimations in the wild at real time, their use in outdoor settings and action sports, \eg alpine skiing, still poses several challenges. Compared to other sports, skiers show very specific poses that are rare in largescale datasets capturing human motion. This applies in particular to out-of-balance and fall situations in alpine ski racing. Furthermore, skiers' poses often show self-occlusions and external occlusions due to snow-spray, gates, or the terrain itself \cite{spoerri}. To overcome some of these challenges, Bachmann \etal \cite{bachmann} created a skiing specific 2D dataset and trained OpenPose \cite{openpose} on this dataset, achieving good results in regular skiing scenarios. However, in injury and fall situations, pose estimation is even more difficult. When the skier is twisted and compressed, self-occlusions, rare and even upside-down poses become more likely, as do occlusions due to snow spray and motion blur. With the goal of developing a deep learning based motion capture tool for the analysis of injury mechanisms, we created an injury specific 2D dataset for alpine skiing, covering 533 manually annotated video frames. To stimulate computer vision research in the injury context of alpine ski racing, the dataset is made publicly available. Running state-of-the-art algorithms on our injury dataset \cite{dcpose, alphapose}, we evaluated their performance during 'regular skiing', as well as in 'out-of-balance' and fall scenarios and identified major difficulties and failure cases. Furthermore, we propose a novel post processing routine, which improves keypoint detection in 'out-of-balance' and fall situations by leveraging information from rotated input frames combined with a kinematic model. The detection results with and without this post processing step were compared as well. 

\begin{figure}[t]
\begin{center}
   \includegraphics[width=0.6\linewidth]{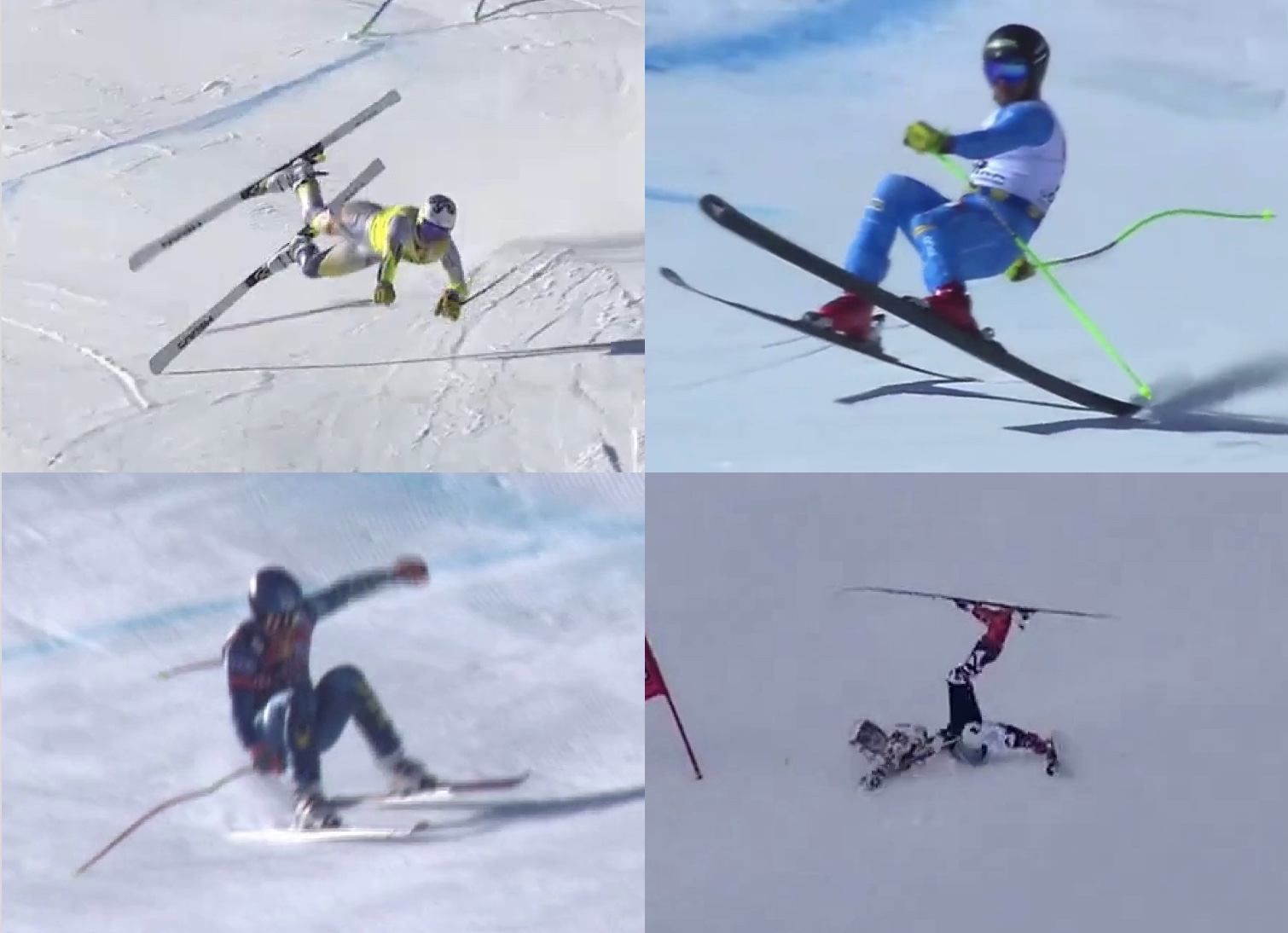}
\end{center}
   \caption{Example images of our injury specific 2D dataset}
\label{fig:dataset1}
\end{figure}


\section{Dataset}
\begin{figure}[b]
\begin{center}
   \includegraphics[width=0.5\linewidth]{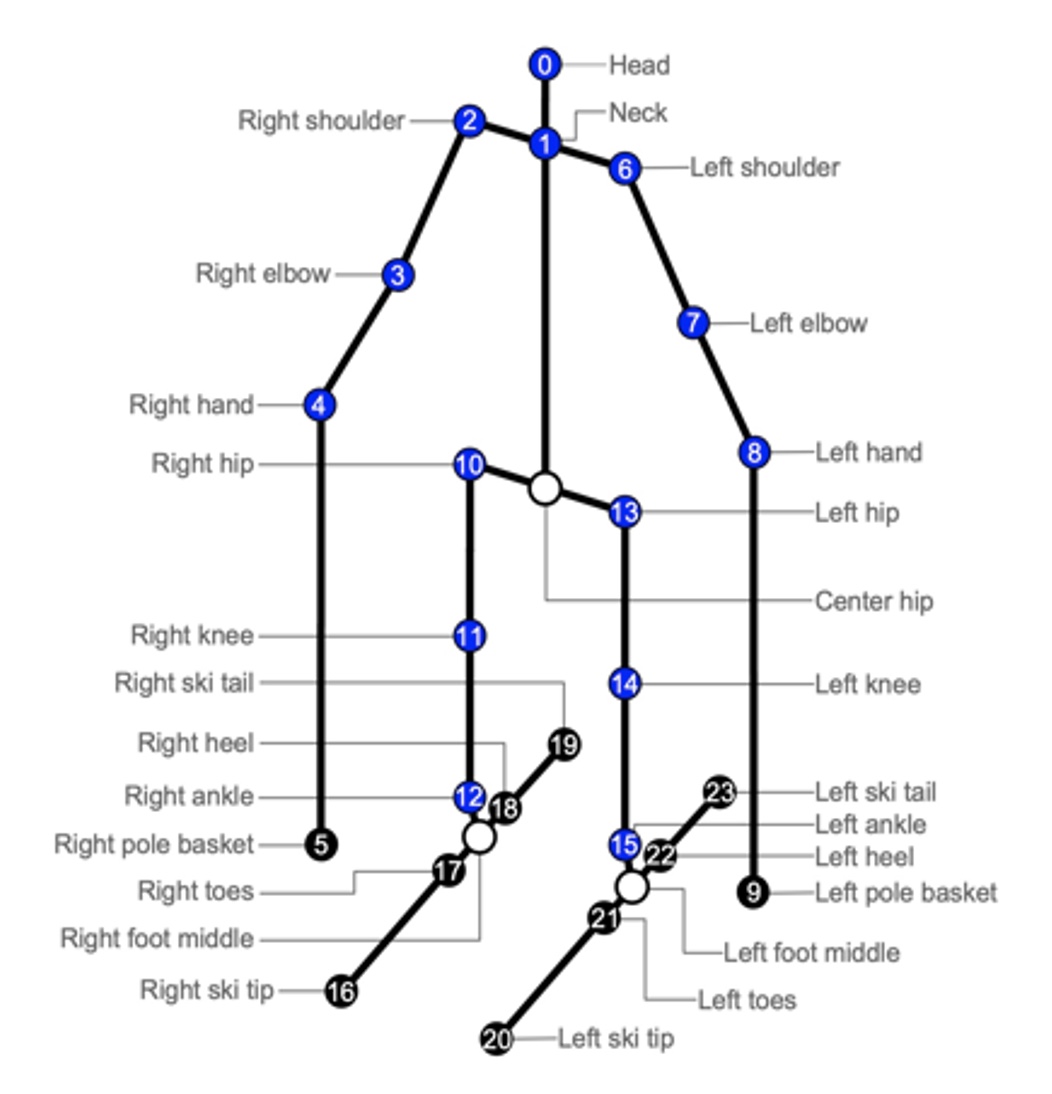}
\end{center}
   \caption{Annotation scheme provided by \cite{bachmann}}
\label{fig:dataset2}
\end{figure}

We created an 'out-of-balance' and fall situation-specific 2D dataset for alpine skiing covering a total of 533 sample images (Figure \ref{fig:dataset1}). Within a large pool of previously collected recordings of ACL injury cases, seven videos were selected for manual annotation. These videos feature (semi-) professional male and female ski racers in each alpine skiing discipline (slalom, giant slalom, super-G and downhill) and cover different video qualities and skier sizes. Only very poor video qualities, not allowing any viable annotation, were discarded. All videos were captured by coaches and/or television broadcasters. Depending on length and framerate, each video was split into 50 to 100 frames in equidistant timesteps and was manually annotated using a custom-built LabVIEW script. Following \cite{bachmann} annotations included 24 keypoints, 16 body joints, 2 pole and 8 ski points. Invisible keypoints were indicated by a visibility flag. The annotation scheme is shown in Figure \ref{fig:dataset2}. At the beginning of each video the skier is still skiing regularly, followed by a phase of losing balance before finally falling. All frames were therefore categorized as ‘regular skiing’, ‘out-of-balance’ or ‘fall’ situations. The performance of the keypoint detection algorithms with and without post processing was evaluated and compared in these categories. Our injury specific 2D dataset is made available online for further research. \url{https://sport1.uibk.ac.at/mz/cv}.

\section{Methods}

Besides occlusions, non-typical poses where the skier is \eg upside-down were identified as a major difficulty for keypoint detection \cite{openpose}. As increasing the rotation augmentation resulted in an overall decrease in accuracy, the authors suggested to run the network using different rotations and keep the poses with higher confidence. Using this idea as a starting point, we developed a novel post processing routine, that combines such rotational information with a kinematic model to increase keypoint detection performance in difficult frames, especially in situations where the subject is orientated horizontally or even upside-down, \eg in fall situations. Therefore, each input video is rotated from $0^{\circ}$ to $360^{\circ}$ in steps of $10^{\circ}$ and then processed by a keypoint detection algorithm, \eg DCPose or AlphaPose. All predicted keypoints are then rotated back to their 'original' location by applying a standard rotation matrix. Thereby, 36 keypoint candidates $p_i = (x_i, y_i, c_i)$ for each keypoint in each frame are obtained, where $x_i$ and $y_i$ refer to the x and y coordinate and $c_i \in [0, 1]$ to the confidence value for each keypoint. As the exact keypoint location is unknown, we apply an $\alpha$-$\beta$-$\gamma$ filter to estimate the keypoint location \cite{abc}. Depending on the location, velocity and acceleration of the keypoint in the current frame $n$, the $\alpha$-$\beta$-$\gamma$ filter provides a kinematic estimation for the keypoints location $p_e$, velocity and acceleration in the next frame $n+1$. Based on this estimate, we select $k$ keypoint candidates, that are closest (smallest spacial distance) to the estimate and calculate the weighted mean $x_m$ and $y_m$ over these 'k-nearest' coordinates $x_i$ and $y_i$. The weights are determined by their confidence values $c_i$. For each frame, we obtain a refined keypoint $p_m = (x_m, y_m)$ which serves as a measurement value for the $\alpha$-$\beta$-$\gamma$ filter. Depending on the parameters $\alpha, \beta, \gamma \in [0,1]$, the filter either tends towards the measurement value, which is in our case the refined keypoint $p_m$, or to the model estimate $p_e$. To optimize these parameters ideally for our injury specific dataset, we conducted a grid search over all parameters $k, \alpha, \beta, \gamma$, which revealed a best overall performance for $k = 12, \alpha = 1, \beta = 1 $ and $\gamma=0$. In this configuration, the quadratic kinematic model underlying the $\alpha$-$\beta$-$\gamma$ filter reduces to a simple linear kinematic model, assuming a keypoint trajectory with zero acceleration and constant velocity. For all results in this study these parameters remained fixed. To evaluate the performance of the proposed post processing routine, we ran DCPose and AlphaPose in their pretrained configuration on our injury specific dataset, with and without applying the post processing. Results were then evaluated using the average precision (AP) and percentage of correct keypoint (PCK) metrics. The PCK threshold was chosen to be 20\% of the torso diameter (PCK@0.2).

\begin{figure*}
\begin{center}
    \includegraphics[width=0.65\linewidth]{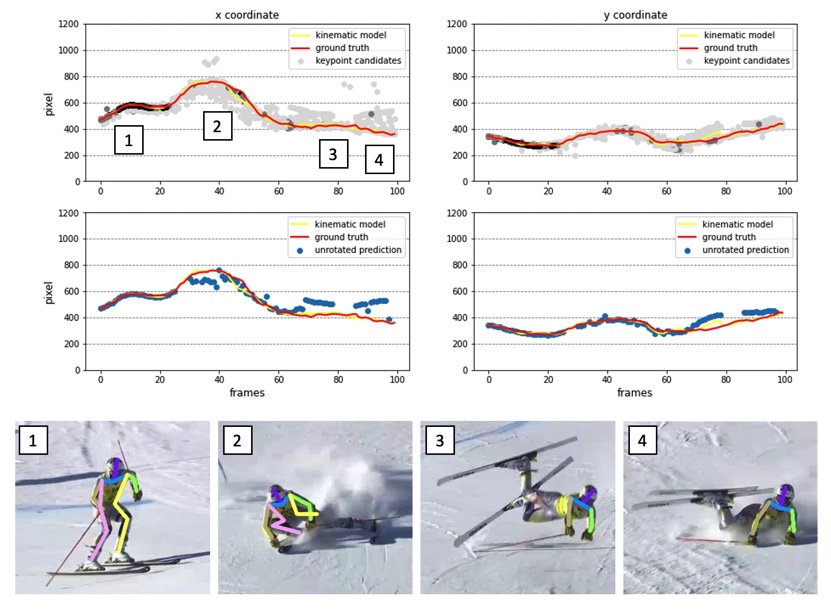}
\end{center}
   \caption{Left ankle coordinates $x$ and $y$ of video n°2 detected by DCPose. The first row shows the keypoint distribution of all rotations. A darker shading indicates a higher confidence value. In the second row, only standard unrotated DCPose detections are graphed in blue dots. Gaps in the sequence of unrotated DCPose detections indicate frames, where the corresponding keypoint was not detected. Ground truth is shown in red, our refined model in yellow. At the bottom we provide some reference frames, overlayed with unrotated DCPose detections above a certain confidence threshold.}
\label{fig:distribution}
\end{figure*}

\section{Results}

\begin{table}[b]
\begin{center}
\begin{tabular}{|l|c|c|c|c|}
\hline
PCK@0.2 & all & reg & oob & fall \\
\hline\hline
DCPose & 0.81 & 0.91 & 0.86 & 0.51 \\
DCPose + PP & 0.87 & 0.93 & 0.88 & 0.72 \\
\hline
AlphaPose & 0.84 & 0.93 & 0.87 & 0.64 \\
AlphaPose + PP & 0.85 & 0.92 & 0.86 & 0.72 \\
\hline
\end{tabular}
\end{center}
\caption{Comparison of DCPose and DCPose + post processing (PP) results by the percentage of correct keypoint (PCK) metric with respect to regular skiing (reg), out-of-balance (oob) and fall frames.}
\label{tbl_pck}
\end{table}

\begin{table}[t]
\begin{center}
\begin{tabular}{|l|c|c|c|c|}
\hline
AP & all & reg & oob & fall \\
\hline\hline
DCPose & 0.61 & 0.78 & 0.70 & 0.16 \\
DCPose + PP & 0.66 & 0.77 & 0.72 & 0.34 \\
\hline
AlphaPose & 0.64 & 0.78 & 0.68 & 0.30 \\
AlphaPose + PP & 0.62 & 0.72 & 0.67 & 0.34 \\
\hline
\end{tabular}
\end{center}
\caption{Comparison of DCPose and DCPose + post processing (PP) results by the average precision (AP) metric with respect to regular skiing (reg), out-of-balance (oob) and fall frames.}
\label{tbl_ap}
\end{table}

As shown in Tables \ref{tbl_pck} and \ref{tbl_ap}, both keypoint detectors showed good results for ‘regular skiing' frames predicting over 9 out of 10 frames correctly. While in ‘out-of-balance’ situations PCK dropped marginally, only half to two third of all ‘fall’ keypoints were detected correctly by the algorithms (DCPose: $0.51$, AlphaPose: $0.64$). Applying our proposed post processing routine to DCPose we could increase the overall PCK by $6 \%$, from 0.81 to 0.87. While just marginal improvements were achieved for regular skiing and out-of-balance situations, fall situations could benefit by $21 \%$, now reaching a PCK of $0.72$. For AlphaPose only a marginal overall increase of 0.01 in keypoint detection performance was observed. Regular skiing predictions as well as out-of-balance prediction performance decreased slightly, while agMU-boain, benefit was observed for fall situations with an increase in PCK of $8 \%$. A similar trend was revealed when comparing the AP metric. DCPose showed an overall increase of $0.05$, with a gain of $0.18$ in fall situations, while regular skiing and out-of-balance situations just increased slightly. For AlphaPose a rather small increase of $0.04$ for fall frames could not compensate a larger decrease of $0.06$ for regular skiing, resulting in a small overall decrease of $0.02$. 

As an example, Figure \ref{fig:distribution} shows the keypoint distribution of the left ankle for subject n°2 for the DCPose algorithm. In the first row, the distribution of all 36 keypoint candidates, ground truth and our post processed model are plotted. The second row shows the unrotated predictions and again ground truth and our model. Some example frames overlayed with the standard DCPose detection are presented for a more detailed discussion. A close look at all keypoint distributions revealed some repeating patterns:

\begin{itemize}
    \item[-] Keypoint distributions were narrow (low variance) for regular skiing situations (Figure \ref{fig:distribution} (1)) and spread towards more difficult out-of-balance and fall frames (Figure \ref{fig:distribution} (4)).
    \item[-] Bimodal distributions were observed when keypoints mismatched, \eg occluded keypoints (Figure \ref{fig:distribution} (2)).
    \item[-] Horizontal or upside-down poses resulted in very poor keypoint detections. For many of these frames no keypoints were detected at all (Figure \ref{fig:distribution} (3)).
\end{itemize}

\section{Discussion}
Running state-of-the-art algorithms, DCPose and AlphaPose, on our injury specific dataset, we observed high performance for ‘regular skiing’ scenarios in PCK@0.2 ($0.9$ and higher) as well as in AP ($0.78$). Accordingly, these accurate predictions translate to a narrow keypoint distribution which was observed in regular skiing and most out-of-balance situations, as shown in Figure \ref{fig:distribution} (1). While for 'regular skiing' frames a high prediction performance was hypothesized, both algorithms still performed well in 'out-of-balance' situations ($PCK@0.2 > 0.86$ and $AP>0.68$). During 'regular skiing', poses are quite similar to motions, that are covered in standard datasets and occlusions are less frequent then in a fall scenario. As predictions do not benefit from rotational information when the athlete is skiing in an upright position, detection results did not increase when applying our post processing routine in this regime. 
In contrast, fall situations pose a much greater challenge for keypoint detection algorithms. As depicted in Figure \ref{fig:distribution} (2) - (4), skiers are likely to get twisted and/or compressed while falling or landing in a horizontal or even upside-down position.  While keypoints in such upside-down positions are rarely detected at all, twisted and crunched positions favor self-occlusions and external occlusions due to snow-spray. This caused a decrease in PCK@0.2 to $0.51$ for DCPose and $0.64$ for AlphaPose, while AP values even dropped to $0.16$ (DCPose) and $0.30$ (AlphaPose). Our results show, that rotated input videos have potential to provide valuable information in such situations. Combined with the kinematic model, prediction results greatly benefited from our post processing routine, leading to an increase in PCK@0.2 of $0.21$ and $0.18$ in AP, respectively. Moreover, analysis of the keypoint distributions revealed, that incorrect non-rotated predictions were either detected in other rotational configurations, or the kinematic model estimated a viable keypoint location if no detection was found at all. While AlphaPose results decreased for regular skiing, we could improve the performance in 'out-of-balance' and fall situations by $8 \%$ in PCK@0.2 and $4 \%$ in AP, which is crucial for injury analysis. To avoid loss of performance in regular skiing situations, our routine should be applied to difficult videos and/or frames, where standard predictions show low performance, \eg in 'out-of-balance' and fall situations. 

\section{Conclusion}
In their pretrained configuration, state-of-the-art algorithms provided good keypoint estimates for 'regular skiing' situations. As their performance decreased substantially in fall scenarios, we developed a post processing routine that specifically targets 'out-of-balance' and fall frames. Our post processing routine improved predictions in such frames greatly. However, results are still too low for viable biomechanical analysis. This also applies for lifting the 2D keypoints to 3D space by a 3D human pose estimation algorithm, which would be the next step for more advanced injury analysis. Using our injury specific dataset for training and refining keypoint detection algorithms could bring further improvements, \eg by a transfer learning approach.

{\small
\bibliographystyle{ieee_fullname}
\bibliography{zwoelferetal_bib}
}

\end{document}